\documentclass[letterpaper,10pt,conference]{ieeeconf}
\IEEEoverridecommandlockouts
\usepackage[T1]{fontenc}
\usepackage{amsmath,amssymb}
\usepackage{graphicx,booktabs,array,multirow}
\usepackage{cite,url,xcolor}
\usepackage{balance}
\newcolumntype{L}[1]{>{\raggedright\arraybackslash}p{#1}}

\newcommand{\setresult}[2]{\expandafter\def\csname result@#1\endcsname{#2}}
\newcommand{\val}[1]{\ifcsname result@#1\endcsname\csname result@#1\endcsname\else\textnormal{--}\fi}

\newcommand{\cfg}[1]{\ifcsname result@#1\endcsname\csname result@#1\endcsname\else\textnormal{[TBD]}\fi}

\makeatletter
\newcommand{\bstcontrol}[1]{\@bsphack\immediate\write\@auxout{\string\citation{#1}}\@esphack}
\makeatother
\title{\LARGE\bf Self-Labeled Preference Transfer for Steerable Robot Manipulation}
\author{Kaiwen Hong$^{1}$, Shaoting Peng$^{1\dagger}$, Xiangyu Zeng$^{1\dagger}$, Pengcheng Wang$^{2}$, Boyuan Tian$^{1}$,\\
Haonan Chen$^{3, 4}$, Yifan Zhu$^{5}$, Yixiao Wang$^{2}$, Katherine Driggs-Campbell$^{1}$
\thanks{$^{1}$University of Illinois Urbana-Champaign}%
\thanks{$^{2}$University of California, Berkeley}%
\thanks{$^{3}$Harvard University}%
\thanks{$^{4}$Stanford University}%
\thanks{$^{5}$University of Illinois Chicago}%
\thanks{$\dagger$ Equal contribution, listed in alphabetical order.}%
}
\begin{document}
\maketitle
\thispagestyle{empty}\pagestyle{empty}
\bstcontrol{IEEEexample:BSTcontrol}
\begin{abstract}

Vision-language-action (VLA) policies can learn to follow execution preferences, such as where to grasp an object or how to move around an obstacle, from demonstrations annotated with preference labels. Providing these
annotations for every new task, however, is costly. We study whether learned preference control can transfer to new tasks whose demonstrations contain the desired behaviors but lack preference annotations. We find that a policy trained with preference annotations can follow preferences on a new task before adaptation but achieves limited task success. Fine-tuning on target demonstrations without these annotations improves task success while weakening preference control. Motivated by this finding, we introduce Self-Labeled Preference Transfer (SPT). SPT jointly trains a shared vision-language backbone for preference-conditioned action prediction and preference recognition on annotated source demonstrations. It then freezes the recognizer to preserve its cross-task preference recognition capability and uses it to self-label target demonstrations for policy fine-tuning. Across five types of execution preferences in simulation, the recognizer achieves 96.4\% accuracy, and SPT reaches 84.8\% preference following and 89.2\% task success, compared with 59.4\% and 88.0\% for naive finetuning. We further extend to multiple preferences, achieving 93.8\% preference following compared with 42.5\% of baselines. In real-world experiments, SPT increases preference following from 52.5\% to 92.5\% relative to naive finetuning across two task transfers. These results show that SPT can support preference
adaptation across tasks and objects while maintaining success, without additional manual preference annotations.

\end{abstract}
\section{Introduction}
Vision-language-action (VLA) models trained on large multi-task datasets can now follow language instructions across scenes, objects, and embodiments~\cite{o2024open,kim2024openvla,black2024pi0,libero,nasiriany2024robocasa,mees2022calvin}. Demonstrations in these datasets usually show one way to complete a task, or follow whatever pattern the operator found natural. In reality, a task has several acceptable executions, and which one is wanted depends on the context and on the user. A mug can be grasped near its rim or near its base; an object can be placed at the center of a tray or in a corner; a path can pass close to an obstacle or give it a wide berth. We call such choices \emph{execution preferences}: they concern \emph{how} a task is performed rather than \emph{what} is accomplished, and the desired choice may depend on the user or context. Standard task-level instructions do not specify this choice, leaving the user without explicit control over it. This paper studies how a VLA can follow a stated execution preference on a new task when its target demonstrations do not contain preference labels.

\begin{figure}[t]
\centering
\includegraphics[width=\linewidth]{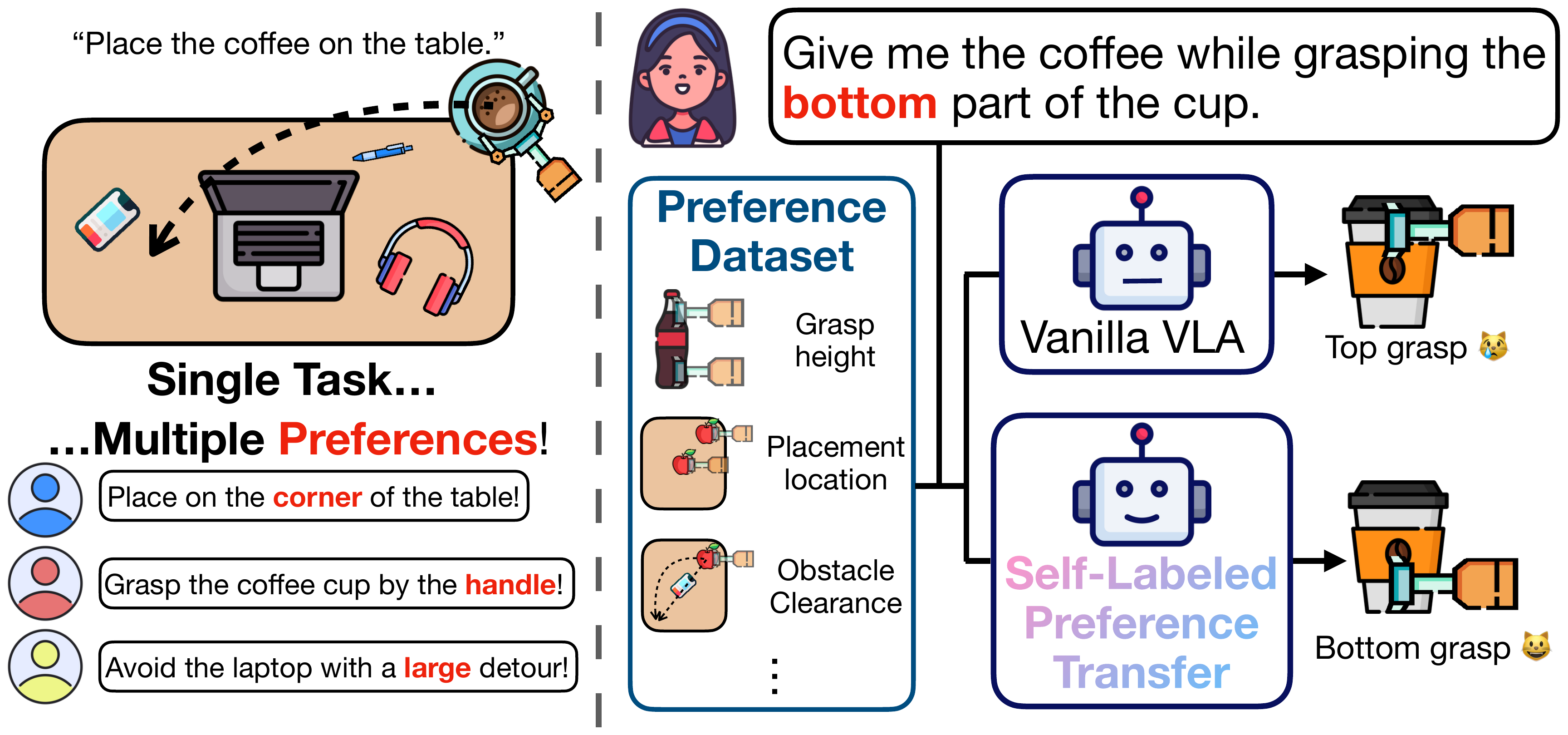}
\vspace{-10pt}
\caption{\textbf{Motivation and overview.} A task can be completed in several valid ways. SPT learns to follow and recognize execution preferences on labeled source tasks. It then labels target demonstrations so that the adapted policy can respond to the same preference commands. Target demonstrations are required; their preference labels are not. The examples illustrate the intended interface, not additional evaluated tasks.}
\label{fig:teaser}
\vspace{-10pt}
\end{figure}

Existing work approaches execution-level control from two directions. One line makes VLAs steerable by enriching training prompts with descriptions of how each demonstration was performed, produced by human annotators or automated pipelines~\cite{steer,finevla,demian,pi07}. These works show that VLAs can follow execution-level instructions once such annotations are available, but a new task still has to be annotated before the policy can be steered on it. The other line is preference learning, which infers a reward or policy from comparisons between executions~\cite{Sadigh2017ActivePL,cpl,zhang2024grape,moletta2026rko}.  
The output of preference learning is a policy aligned to a user's preference (often through ranking executions~\cite{Peng2025UUPL}) rather than one that switches on command to a named style.
Preference learning is typically used to adapt a policy toward preferred behavior on a particular task, rather than to provide command-conditioned switching among named execution preferences across tasks. Neither line asks whether preference control learned on labeled source tasks can be preserved when adapting to a new task whose demonstrations lack preference labels. This setting introduces two challenges. First, the source supervision must teach preference semantics that transfer across tasks, so that the behavior shown in a target demonstration can still be recognized. Second, the policy must adapt to the target task without losing its response to the preference command. Finetuning on target demonstrations with the task instruction alone provides no signal tying the demonstrated motion to a preference, so the control learned on the source can be weakened during adaptation. 

We study preference transfer without target preference annotations. We assume preference-labeled demonstrations for a set of source tasks and demonstrations with varied execution behaviors, but no preference labels, for a set of target tasks. Our approach, \textbf{Self-Labeled Preference Transfer (SPT)}, addresses this setting in two stages. In \emph{Stage~A}, a shared vision-language backbone learns both to execute a stated preference and to recognize the preference expressed in a demonstration. In \emph{Stage~B}, we freeze the Stage-A recognizer and use it to label the target demonstrations, then finetune a trainable copy of the policy on the target actions paired with these inferred preference prompts. These labels associate each target behavior with the corresponding preference command during adaptation. At deployment, only the policy runs, and the user selects a behavior using the same preference vocabulary learned on the source tasks.

We evaluate SPT in simulation and on a real robot. In RoboTwin~2.0~\cite{robotwin}, we synthesize pick-and-place demonstrations along five preference axes and transfer to held-out object and receptacle combinations. On a real robot experiment, we transfer across task types: from pick-and-place to pushing and from air hockey to billiards. The Stage-A recognizer labels target demonstrations with 96.4\% average accuracy, and SPT raises average preference following from 59.4\% to 83.8\% in simulation and from 52.5\% to 92.5\% on the robot, with similar average task success.

Ablations on placement and contact height isolate the roles of source preference transfer and target self-labeling. Before target adaptation, the Stage-A policy reaches 80\% and 88\% preference following on the two target tasks, but with lower target-task success. Finetuning the same Stage-A checkpoint with bare target instructions reduces preference following to 52\% and 50\%, whereas SPT retains 83\% and 90\% while adapting to the target tasks. Shuffling the inferred target labels gives similarly poor preference following, indicating that the benefit depends on correctly associating each target behavior with its preference rather than on adding preference text alone.


In summary, this paper makes three contributions. (1)~We formulate transfer of a named execution-preference vocabulary to target tasks whose demonstrations lack preference annotations. (2)~We introduce SPT, which jointly learns preference-conditioned action prediction and preference recognition, then reuses a frozen recognizer to label target demonstrations before policy adaptation. (3) We evaluate this approach across five simulated preference axes and two real-robot task transfers, with controls for source initialization and target label assignment, comparisons to target-labeled reference conditions, and an extension to two simultaneous preferences.
\section{Related Work}
\label{sec:related}
\textbf{Steerable VLAs.}
Large multi-task datasets have produced policies that condition on language and generalize across tasks, scenes, objects, and embodiments~\cite{o2024open,kim2024openvla,black2024pi0}, usually by attaching an action head to a pretrained VLM and finetuning on demonstrations~\cite{kim2024openvla,kim2025oft}. Their language conditioning is not always robust~\cite{liberoplus}, and the goal-level instructions in most datasets say nothing about execution. A recent line of work addresses this by writing execution details into the training language. STEER grounds modular skills in dense attribute annotations~\cite{steer}; DeMiAn and FineVLA re-annotate existing datasets with fine-grained process descriptions, through a VLM pipeline in one case and human-verified labels plus a trained robotics VLM annotator in the other~\cite{demian,finevla}; and $\pi_{0.7}$ trains a generalist on prompts that carry strategy and metadata alongside the goal~\cite{pi07}. These works establish that a VLA follows how-level instructions once its training data contains them, and each obtains that data by annotating it. SPT differs in what it asks and where the labels come from. We assume a labeled source set and ask whether preference recognition learned jointly with the source policy can transfer to a new task, so that its demonstrations can be labeled without additional preference annotation and the adapted policy can still respond to the source vocabulary.

\textbf{Preference learning and personalization.}
Preference learning typically learns rewards or policies from comparisons between executions~\cite{Sadigh2017ActivePL,cpl,Peng2025UUPL}. In manipulation, GRAPE, APO, and RKO apply preference optimization to VLAs and diffusion policies~\cite{zhang2024grape,xia2025hapo,moletta2026rko}, and Preference VLM, VARP, and VLP replace human comparisons with VLM judgments, the last training an annotator that transfers to downstream tasks~\cite{ghosh2025prefvlm,singh2025varp,vlp}. Personalization work asks how a robot should choose among several valid behaviors for a given user: COMMET resolves human-induced conflicts by selecting among options according to stored user preferences~\cite{li2025commethumaninducedconflictsmobile}, TacStyle has an LLM pick a point in a learned style latent from a handful of labels~\cite{tacstyle}, and Freeform Preference Learning trains a policy that responds to several language-defined preference axes~\cite{fpl}. Our setting shares the premise that the user, not the task, decides the execution, but the supervision differs. A label such as ``grasp low'' describes a style without ranking it above ``grasp high''; both are correct when requested. Rather than aligning a policy separately on each task, we ask whether preference following learned on labeled source tasks can be transferred to target tasks whose demonstrations lack preference labels.

\textbf{Self-labeling and auxiliary supervision in VLAs.}
Using a model's own predictions as labels for unlabeled data is standard in semi-supervised learning~\cite{fixmatch}, and foundation models have been used to label robot data with task instructions~\cite{blank2024nils} and process descriptions~\cite{demian}. In VLAs, auxiliary language objectives are usually added to improve representations: LAP co-trains action prediction with visual question answering~\cite{zha2026lap}, SG-VLA adds spatial grounding~\cite{tu2026sgvla}, and TwinBrainVLA keeps a generalist pathway beside the action pathway~\cite{yu2026twinbrainvla}. SPT combines these two ideas but uses them differently. The recognition objective is not only an auxiliary training signal: the resulting recognizer is reused to label demonstrations from a different task.

\begin{figure*}[!th]
\centering
\includegraphics[width=0.95\textwidth]{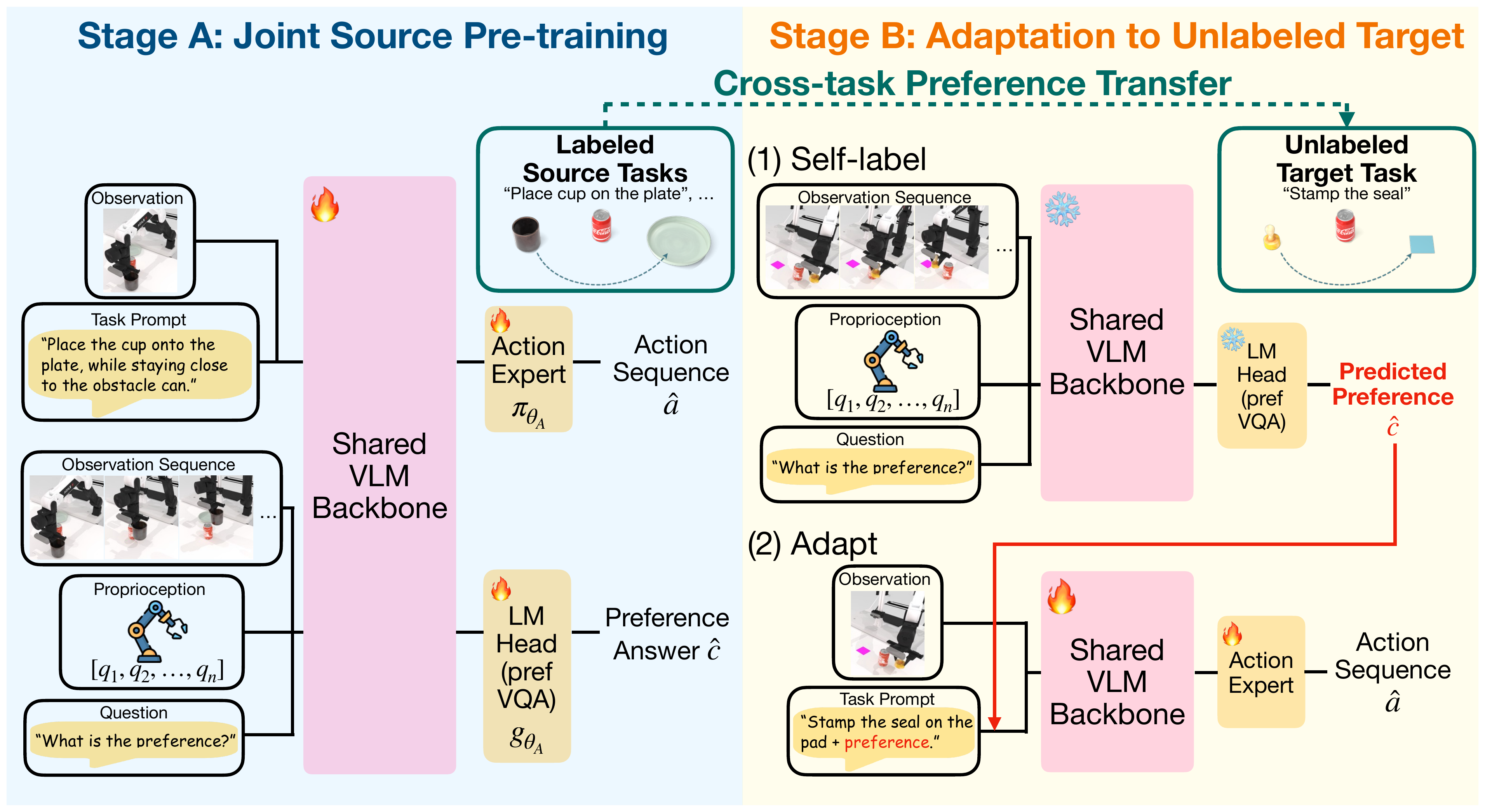}
\vspace{-5pt}
\caption{\textbf{Self-Labeled Preference Transfer.} Stage~A trains a shared vision-language backbone for preference-conditioned action prediction and preference recognition. The recognition branch answers a preference question from a short demonstration clip and proprioceptive observations. Stage~B has two steps: a frozen copy of the source recognizer labels each target demonstration, and a trainable policy copy adapts to the resulting preference-conditioned prompts. At deployment the user gives a preference command and only the action policy runs.}
\vspace{-10pt}
\label{fig:method}
\end{figure*}

\section{Problem Formulation}
\label{sec:formulation}
We explore learning from demonstrations, where style and preferences are considered through a \emph{preference axis} that is only available on the source data and not on the final target domain. The preference axis $\mathcal{C}$ is a finite vocabulary set that captures preferred behaviors that modify aspects of the task (e.g., low grasp contact height or high placement location).  
The user selects a behavior by naming a value from this vocabulary.
Suppose we are given the following data from an initial source set of tasks:
\begin{equation}
\mathcal D_{\mathcal S}=\{(\xi_{\mathcal S}^{(i)},\ell_{\mathcal S}^{(i)},c^{(i)})\},
\end{equation}
where $\xi_\mathcal{S}$ are sample trajectories containing sensor observations, proprioceptive states, and actions, $\{(o_t,s_t,a_t)\}_{t=1}^{T}$. 
$\ell_\mathcal{S}$ is the language instruction that guided the operator for the source demonstration. 
$c$ is the relevant label for the style of the demonstration. 
$i$ denotes the sample index.



Given $\mathcal{D}_\mathcal{S}$, we would like to transfer skills and preferences to new target tasks that share the same preference vocabulary $\mathcal C$. We assume we are given demonstrations from the target domain (following typical fine tuning frameworks), but do not assume access to preference labels: 
\begin{equation}
\mathcal D_{\mathcal T}=\{(\xi_{\mathcal T}^{(j)},\ell_{\mathcal T}^{(j)})\}.
\end{equation}
where variables follow structure from the source domain, but are specific to the target task space $\mathcal{T}$.

The goal is a policy $\pi_\theta(a \mid o, s, \ell, c)$ that completes a
target task in the style named by $c \in C$. We also learn a preference
recognizer that predicts the preference expressed by an observed
demonstration; this recognizer is later frozen and used to label target
demonstrations in Stage~B. Task completion and preference following are
scored separately.

Two assumptions scope the setting. First, the target demonstrations must contain the requested styles: we remove their labels, not the behaviors, and self-labeling cannot produce a behavior that appears in neither the source nor the target data. Second, the vocabulary is fixed, so a value outside $\mathcal C$ is out of scope. The single-axis experiments use one binary axis at a time; Section~\ref{sec:multi} covers two axes in one command.

\section{Methods}
\label{sec:methods}
SPT has two stages (Fig.~\ref{fig:method}). Stage~A jointly trains a model on the labeled source tasks to act on a stated preference and to recognize the preference shown in a demonstration. Stage~B freezes the Stage-A recognizer to label the target demonstrations, then adapts a trainable copy of the Stage-A policy on the self-labeled target data.

\textbf{Architecture.}
The action branch and the recognition branch share one pretrained vision-language backbone. The action head predicts a chunk of actions from the current observation and the instruction, to which we append ``Preference: $w(c)$'', where $w(c)$ is a fixed phrase such as ``grasp low''. The recognition branch reuses the VLM's language head instead of adding a classifier. It receives a short demonstration clip, proprioceptive information, and a question about the preference, and for the binary axes used here it answers with a single token. 

\textbf{Stage~A: joint source training.}
Each training batch mixes action samples and recognition samples. An action sample pairs an observation and a preference-conditioned instruction with the demonstrated action chunk, and contributes a regression loss $\mathcal L_{\mathrm{act}}$. A recognition sample pairs a demonstration clip with its question and answer, and contributes a cross-entropy loss $\mathcal L_{\mathrm{inf}}$. The objective is
\begin{equation}
\mathcal L_A=\mathcal L_{\mathrm{act}}+\lambda\mathcal L_{\mathrm{inf}},
\label{eq:stageA}
\end{equation}
where $\lambda$ weights the recognition loss. The result is a source model $M_A=(\pi_{\theta_A},g_{\theta_A})$: the policy $\pi_{\theta_A}$ acts on a stated preference, and the recognizer $g_{\theta_A}$ predicts the preference expressed in a demonstration.

\textbf{Stage~B: labeling and adapting on the target.}
We first freeze a copy of $g_{\theta_A}$ and let it label every target demonstration:
\begin{equation}
\hat c^{(j)}=\arg\max_{c\in\mathcal C} g_{\theta_A}(c\mid o^{(j)},s^{(j)},q),
\label{eq:pseudolabel}
\end{equation}
where $o^{(j)}$ denotes the sampled observation clip, $s^{(j)}$ its corresponding proprioceptive states, and $q$ the fixed question for the preference axis. The predicted label is cached and appended to the demonstration's instruction. Target ground-truth preference labels are never provided to the recognizer or the adapting policy. 

We then initialize a trainable policy from the Stage-A policy $\pi_{\theta_A}$ and adapt it on the self-labeled target demonstrations with the action loss alone:
\begin{equation}
\mathcal L_B=\mathcal L_{\mathrm{act}}^{\mathcal T}\big(w(\hat c)\big).
\label{eq:stageB}
\end{equation}
The recognizer remains frozen while the policy is adapted, so the cached target labels remain fixed throughout Stage~B. No recognition loss is applied during target adaptation.

\textbf{Deployment.}
The user gives a target instruction and a preference from the source vocabulary, and the adapted policy acts on the combined prompt. The recognition branch is not invoked at deployment, so SPT requires no additional model call at inference time. For real-world rollouts, we use discreteRTC for action execution~\cite{wang2026discretertc}.

\textbf{Multiple preferences.}
For $K$ axes the command is $\mathbf c=(c_1,\ldots,c_K)$ and every phrase $w(c_k)$ is appended to the instruction. The same language head answers one question per axis, and the recognition loss becomes
\begin{equation}
\mathcal L_{\mathrm{inf}}^{\mathrm{multi}}=-\frac{1}{K}\sum_{k=1}^{K}\log g_\theta(c_k\mid\xi,q_k).
\label{eq:multi}
\end{equation}
In Stage~B the labeler predicts each component before adaptation. Section~\ref{sec:multi} evaluates this with two axes.

\begin{figure*}[t]
\centering
\includegraphics[width=\textwidth]{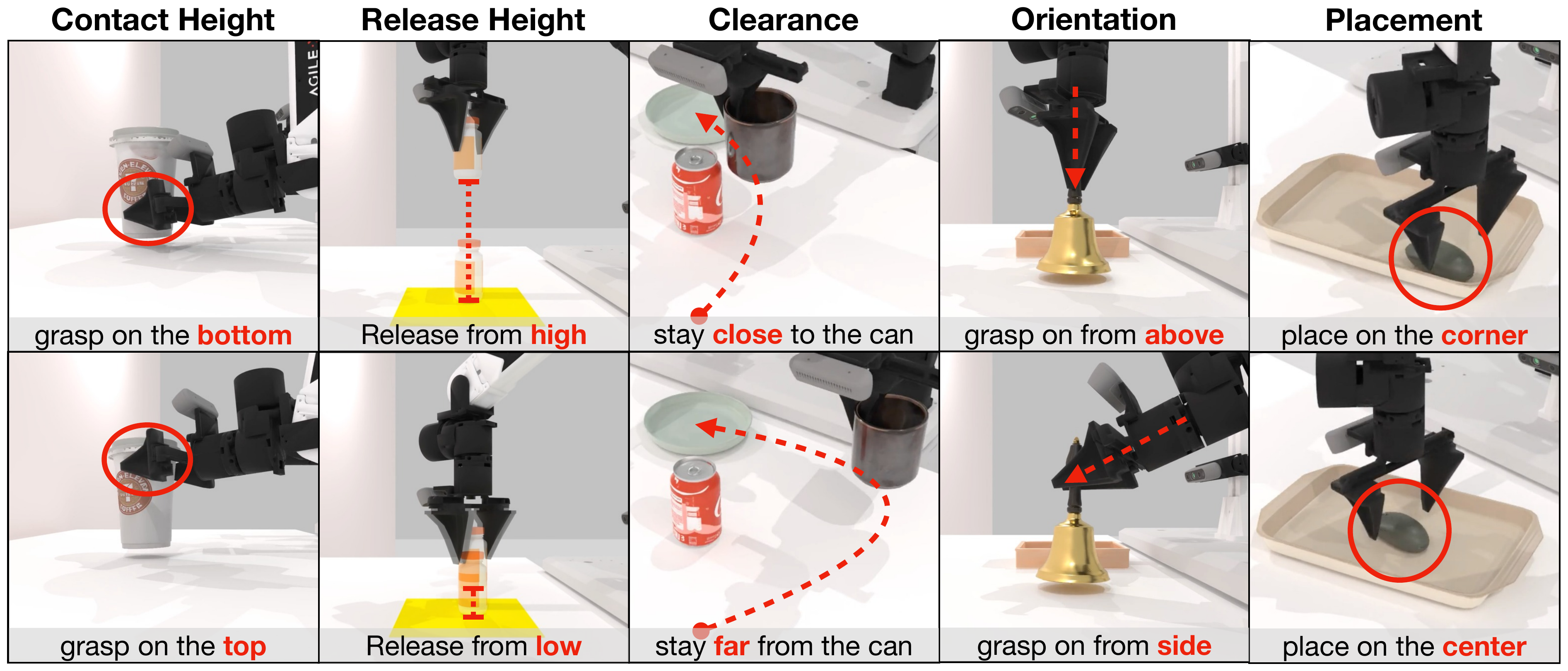}
\vspace{-15pt}
\caption{\textbf{Illustration of preferences in simulation.} The base instruction is ``Pick up A and place it on B'' with different objects and receptacles in the source and target sets. Columns show the five preference axes and rows show their two values. The single-axis experiments evaluate each preference type separately; Section~V-E combines release height and placement preferences.}
\label{fig:sim}
\end{figure*}

\begin{figure*}[t]
\centering
\includegraphics[width=0.99\textwidth]{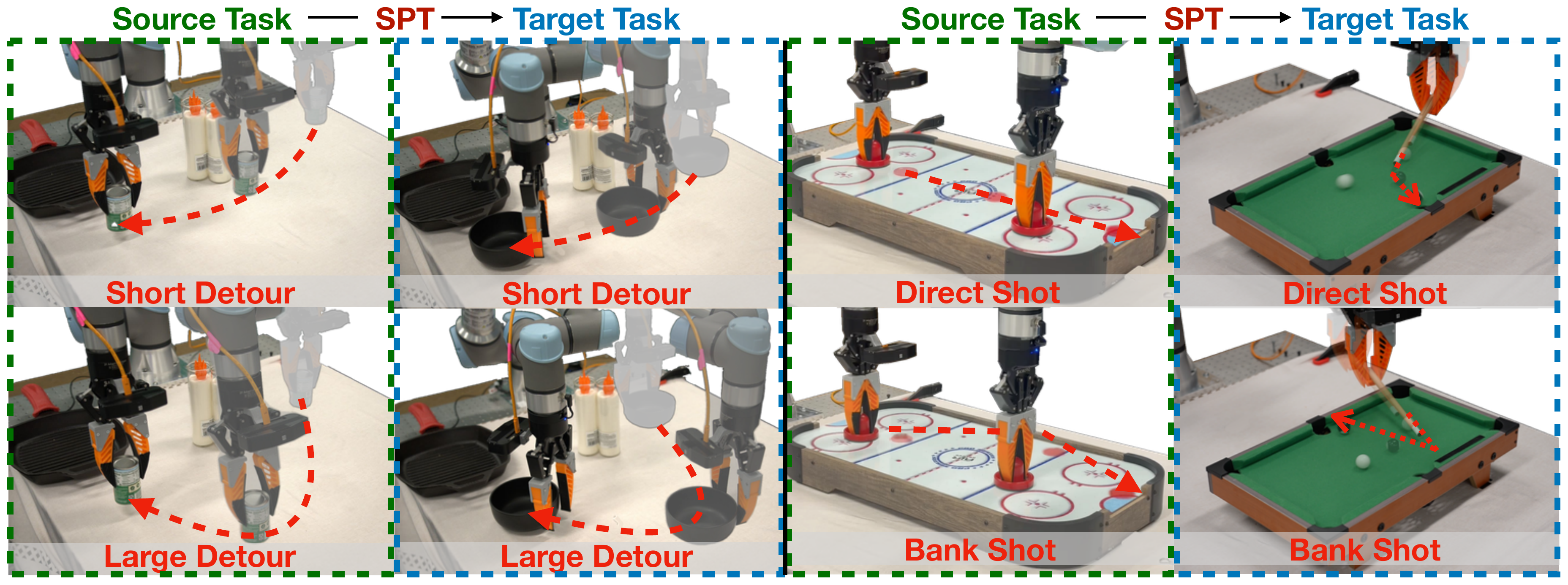}
\vspace{-5pt}
\caption{\textbf{Real-robot setup.} Left: pick-and-place (source) and pushing (target) with two obstacle-clearance preferences. Right: air hockey (source) and billiards (target) with direct and bank-shot strategies. Dashed paths show the requested behaviors.}
\vspace{-10pt}
\label{fig:real}
\end{figure*}

\section{Experiments}
\label{sec:experiments}

We evaluate SPT through four sets of experiments. \textbf{C1} tests whether preference recognition learned in Stage~A transfers to target demonstrations. \textbf{C2} evaluates the complete SPT pipeline for preference-controlled target adaptation in simulation and on a real robot. \textbf{C3} ablates the two-stage design to examine how Stage~A transfer, target adaptation, and target preference labeling affect task success and preference following. \textbf{C4} extends the setting to two preferences specified in the same command.

\subsection{Experimental Setup and Evaluation}
\label{sec:setup}

\textbf{Simulation.}
We generate preference-varied demonstrations with RoboTwin~2.0 expert scripts~\cite{robotwin} along five axes: grasp contact height (low or high on the object), release height (low or high), obstacle clearance (narrow or wide detour), grasp orientation (side or top), and placement location (center or corner). Fig.~\ref{fig:sim} illustrates the preference axes with different objects, note that all preferences apply to all objects. The source set contains eight tasks per axis, with 100 demonstrations per preference value per task. Each axis has one held-out target task with 100 demonstrations, balanced at 50 per preference value. 

\textbf{Real robot.}
We evaluate two preference transfers between different task types (Fig.~\ref{fig:real}) in real world with a UR5e robot arm. For obstacle clearance, we transfer from pick-and-place around an obstacle to bowl pushing, with near and far path preferences. For strike strategy, we transfer from air hockey to billiards, with direct and bank-shot preferences. The source and target tasks differ in their objects and contact dynamics, while sharing the same goal and the corresponding preference axes. Each source and target set contains 100 demonstrations, with 50 per preference value. Target preference annotations are hidden during adaptation. Pick-and-place and bowl-pushing demonstrations are collected using FastUMI~\cite{chi2024umi, zhaxizhuoma2025fastumi}. Each method is evaluated for ten trials per preference command.

\textbf{Prompts.}
The action prompt consists of the task instruction and, when a preference label is available, the suffix ``Preference: $w(c)$''. For example: ``Pick up the cup and place it on the plate while grasping it low.'' Source policy training uses the ground-truth preference label. During target adaptation, SPT appends the recognizer's inferred label, while Naive FT uses the bare instruction. The recognizer receives a fixed binary question for each axis, such as ``Did the robot grasp the object low or high?'', and predicts the preference label as a single-token answer.

\textbf{Metrics.}
\emph{Recognition accuracy} is the fraction of target demonstrations whose predicted preference matches the ground truth. For SPT, these annotations are used only for evaluation and are not exposed during target adaptation. Chance is 50\% on every axis. \emph{Preference following} is the fraction of all assigned rollouts that satisfy the complete requested preference command. \emph{Task success} measures task completion over the same rollouts. For the multi-preference experiment in Section~\ref{sec:multi}, we also report \emph{joint success}, the fraction that both complete the task and follow the command. A rollout can succeed a task but not follow a preference, then it contributes to \emph{Task success} but not to \emph{Preference following} and \emph{joint success}. 


Thresholds are calibrated from source demonstrations and fixed before target evaluation. For clearance, we select the arm with the larger XY path length and use the trajectory segment between 25\% and 85\% of the rollout. Clearance is the maximum perpendicular XY deviation from the line joining the two endpoints of this segment. Table~\ref{tab:definitions} summarizes the rollout statistic and fixed threshold for each simulated preference axis.

\begin{table}[t]
\caption{Simulation preference metrics and fixed source-calibrated thresholds.}
\label{tab:definitions}
\vspace{-5pt}
\centering\small
\begin{tabular}{@{}L{0.26\linewidth}L{0.67\linewidth}@{}}
\toprule
Axis & Demonstration control and rollout statistic\\
\midrule
Contact height & Grasp at 20--30\% or 70--80\% of object height. EE vertical offset from the object origin at grasp; threshold 6.83\,cm.\\
Release height & Low or high release. Object vertical displacement from grasp to immediately before release; threshold 12.45\,cm.\\
Clearance & Direct path or 20--25\,cm detour. Maximum XY deviation from the 25\%--85\% trajectory chord; threshold 15.01\,cm.\\
Orientation & $0^\circ$ or $90^\circ$ grasp. Left-wrist orientation feature at grasp; threshold $41.22^\circ$.\\
Placement & Center or corner target. EE XY distance to the receptacle's initial actor position at release; threshold 5.36\,cm.\\
\bottomrule
\end{tabular}
\vspace{-5pt}
\end{table}

\textbf{Baseline.}
\emph{Naive FT} starts from the same manipulation-pretrained checkpoint as SPT and follows the same source-then-target optimization schedule. It first trains for 10000 steps on preference-labeled source demonstrations using only the action loss, then adapts for 1500 steps on the 100 target demonstrations using bare task instructions. It uses neither the recognition objective nor inferred target labels. The baseline therefore matches the amount of source and target optimization, while Section~\ref{sec:controls} provides matched controls that isolate the role of self-labeling during target adaptation.

\textbf{Implementation.}
We initialize the policy from a StarVLA-$\alpha$ RoboTwin~2.0 checkpoint~\cite{starvlaalpha}, which uses Qwen3-VL-4B-Instruct~\cite{qwen} as the vision-language backbone and a residual MLP head for continuous action-chunk prediction. In simulation, the policy receives three $224\times224$ RGB views and the instruction, without proprioceptive inputs. The action head predicts 50-step chunks of 14-dimensional joint actions using an L1 loss. The recognizer receives three head-camera frames and the active arm's 9-D EE state at each frame, comprising position and a 6-D rotation representation. A learned projector converts each state into a soft token for the shared backbone.

Stage~A jointly trains action prediction and single-token preference recognition using $\lambda=0.5$ in Eq.~\ref{eq:stageA}. We use an 80/20 source train/validation split, 10000 source-training steps, and 1500 target-adaptation steps, with an effective batch size of 64. Stage~B uses all 100 target demonstrations and the action loss only. The labeler remains frozen throughout adaptation, and matched controls use the same optimization schedule.

For the policy results in Table~\ref{tab:sim_main}, we use three independent training seeds and evaluate 50 rollouts per preference command for each seed. We first average across the two preference commands for each axis and then across training seeds.

\subsection{C1: Target Preference Recognition}
\label{sec:exp-rec}

We evaluate the source-trained recognizer on the 100 target demonstrations for each preference axis before target adaptation. As an off-the-shelf vision-only reference, we use Qwen3-VL-4B-Instruct without source preference training. It receives the same three head-camera frames and binary question and scores the two candidate answers using black-image prior calibration. SPT additionally receives EE-state tokens and starts from a manipulation-pretrained checkpoint. This comparison therefore does not isolate the effect of source preference training. The recognition columns in Table~\ref{tab:sim_main} report label accuracy; the remaining columns report the policy results discussed in Section~\ref{sec:exp-main}.


\begin{table}[t]
\caption{RoboTwin simulation results (\%). Preference recognition accuracy is evaluated on target demonstrations; task success and preference following are evaluated on policy rollouts.}
\vspace{-5pt}
\label{tab:sim_main}
\label{tab:sim_recognition}
\centering\small
\setlength{\tabcolsep}{0pt}
\begin{tabular*}{\columnwidth}{@{\extracolsep{\fill}}lcccccc@{}}
\toprule
& \multicolumn{2}{c}{\shortstack{Preference\\Recognition Acc.}}
& \multicolumn{2}{c}{\shortstack{Task\\Success}}
& \multicolumn{2}{c}{\shortstack{Preference\\Following}}\\
\cmidrule(lr){2-3}\cmidrule(lr){4-5}\cmidrule(lr){6-7}
Preference axis & Zero-shot & SPT & Naive FT & SPT & Naive FT & SPT\\
\midrule
Contact height & 51 & \textbf{100} & 92 & \textbf{95} & 59 & \textbf{90}\\
Release height & 46 & \textbf{100} & 89 & \textbf{91} & 53 & \textbf{81}\\
Clearance & 53 & \textbf{92} & \textbf{85} & 82 & 57 & \textbf{75}\\
Orientation & 50 & \textbf{100} & \textbf{94} & 92 & 61 & \textbf{90}\\
Placement & 59 & \textbf{90} & 80 & \textbf{86} & 67 & \textbf{83}\\
\midrule
Average & 51.8 & \textbf{96.4} & 88.0 & \textbf{89.2} & 59.4 & \textbf{83.8}\\
\bottomrule
\end{tabular*}
\vspace{-10pt}
\end{table}

The Stage-A recognizer achieves 96.4\% average accuracy on target demonstrations. It reaches 100\% on contact height, release height, and grasp orientation, 92\% on obstacle clearance, and 90\% on placement location. The off-the-shelf zero-shot VLM reference achieves 51.8\% on average. Thus, the recognizer learned on the source tasks generalizes well enough to provide high-quality labels for target adaptation across all five preference axes.

\emph{Summary.} The Stage-A recognizer transfers to the target demonstrations with 96.4\% average accuracy and at least 90\% accuracy on every axis. These predictions provide the preference labels used for target adaptation in Stage~B, without exposing the model to target preference annotations.



\subsection{C2: Preference-Controlled Target Adaptation}
\label{sec:exp-main}

\textbf{Simulation.}
Table~\ref{tab:sim_main} compares Naive FT and SPT across all five axes. SPT improves average preference following from 59.4\% to 83.8\%, a gain of 24.4 percentage points. Following improves on every axis, with gains ranging from 16 points for placement to 31 points for contact height. Average task success is 89.2\% for SPT and 88.0\% for Naive FT. Task success is not uniformly higher: SPT is lower than Naive FT on clearance and orientation, despite improving preference following. Section~\ref{sec:controls} shows that the gain is not explained by the Stage-A initialization alone: finetuning the same Stage-A policy on the same target demonstrations without self-labels substantially weakens preference control.



\textbf{Preference following on the real robot with task transfer.}
SPT follows the requested preference in 37 of 40 trials, compared with 21 of 40 for Naive FT (Table~\ref{tab:real_main}). It follows both clearance requests in all 20 pushing trials and the direct and bank-shot preferences in 17 of 20 billiards trials. Task success is 38 of 40 for SPT and 36 of 40 for Naive FT. These results show that clearance preferences learned from pick-and-place transfer to bowl pushing, while strike preferences learned from air hockey transfer to billiards, without preference annotations on the target demonstrations.


\begin{table}[t]
\caption{Real-robot results (\%), aggregated over both preference commands for each target task (20 trials per task).}
\vspace{-5pt}
\label{tab:real_main}
\centering\small
\setlength{\tabcolsep}{0pt}
\begin{tabular*}{\columnwidth}{@{\extracolsep{\fill}}lcccc@{}}
\toprule
& \multicolumn{2}{c}{Task Success}
& \multicolumn{2}{c}{Preference Following}\\
\cmidrule(lr){2-3}\cmidrule(lr){4-5}
Target task & Naive FT & SPT & Naive FT & SPT\\
\midrule
Pushing   & 95 & \textbf{100} & 55 & \textbf{100}\\
Billiards & 85 & \textbf{90}  & 50 & \textbf{85}\\
\bottomrule
\end{tabular*}
\vspace{-10pt}
\end{table}

\emph{Summary.} SPT improves preference following over Naive FT by 24.4 percentage points in simulation and 40 points on the real robot, with similar average task success. The improvement holds across all five simulated axes and both real-robot task transfers.

\subsection{C3: Ablation Studies: Why Are Both Stages Necessary?}
\label{sec:controls}

\textbf{Source initialization and target adaptation.}
Table~\ref{tab:controls} studies placement location and contact height. \emph{Source-only} evaluates the joint Stage~A policy directly on the target tasks without target adaptation. All matched adaptation controls start from the same joint Stage~A checkpoint as SPT and use the same 100 target demonstrations and 1500-step adaptation schedule. \emph{Same-init plain FT} adapts using bare task instructions. \emph{random labels} randomly permutes the inferred preference labels across target demonstrations while preserving the label histogram. \emph{Zero-shot VLM} uses labels predicted by the off-the-shelf VLM from Section~\ref{sec:exp-rec}. SPT instead uses labels from the Stage~A recognizer. Naive FT is the full-pipeline baseline defined in Section~\ref{sec:setup}.

\emph{Few-label (10 GT)} adapts on ten target demonstrations with ground-truth preference labels; the remaining 90 demonstrations are not used in this run. \emph{Oracle GT} uses ground-truth preference labels for all 100 target demonstrations. Both start from the same joint Stage~A checkpoint. These provide references at two levels of target preference supervision, although Few-label also uses less target action data than the full-data methods.



\begin{table}[t]
\caption{Ablations of the two-stage design (\%) on placement location and grasp contact height. Bold marks the best result among methods without target preference annotations; ties are also bold. Supervised references are shown below the midrule.}
\vspace{-5pt}
\label{tab:controls}
\centering\small
\setlength{\tabcolsep}{4.5pt}
\begin{tabular}{@{}lcccc@{}}
\toprule
&\multicolumn{2}{c}{Task Success}
&\multicolumn{2}{c}{Preference Following}\\
\cmidrule(lr){2-3}\cmidrule(lr){4-5}
Method & Placement & Contact & Placement & Contact\\
\midrule
Naive FT &80&92&67&59\\
Source-only &67&85&80&88\\
Same-init plain FT &69&\textbf{95}&52&50\\
random labels &66&92&50&55\\
Zero-shot VLM &70&93&60&50\\
SPT (Ours) &\textbf{86}&\textbf{95}&\textbf{83}&\textbf{90}\\
\midrule
Few-label GT &81&91&75&86\\
Oracle GT &84&98&98&96\\
\bottomrule
\end{tabular}
\vspace{-10pt}
\end{table}

\textbf{Effect of target adaptation.}
Source-only already exhibits preference control on the target tasks, reaching 80\% following on placement and 88\% on contact height, but achieves task success of only 67\% and 85\%. Adapting the same Stage~A checkpoint using bare target instructions yields 69\% and 95\% task success, while preference following drops to 52\% and 50\%. SPT instead reaches 86\% and 95\% task success while retaining 83\% and 90\% preference following. These results show that Stage~A transfers preference-conditioned behavior, while self-labeled target adaptation improves adaptation to the target tasks without erasing that control.

\textbf{Effect of label assignment.}
random labels yields 50\% preference following on placement and 55\% on contact height, while labels from the zero-shot VLM yield 60\% and 50\%. SPT reaches 83\% and 90\%, respectively. Because these conditions use the same Stage~A initialization, target demonstrations, and adaptation schedule, the results indicate that target adaptation depends on labels that reliably identify the preference expressed in each demonstration.

\textbf{Ground-truth supervision.}
Oracle GT reaches 98\% preference following on placement and 96\% on contact height, compared with 83\% and 90\% for SPT. This leaves gaps of 15 and six percentage points to the fully supervised reference. Few-label (10 GT), which uses ground-truth preference labels for only ten target demonstrations, reaches 75\% and 86\%. Because it also uses substantially less target action data, we treat it as a low-annotation reference rather than a matched comparison of label quality.


\emph{Summary.} Stage~A transfers preference-conditioned behavior to the target tasks, but the source-only policy remains less adapted to the target. Fine-tuning on bare instructions weakens preference control, while Stage~B uses inferred labels to adapt without losing that control. random and zero-shot labels perform substantially worse, indicating that effective target adaptation requires reliable preference labels. SPT therefore approaches the fully supervised reference without target preference annotations.

\subsection{C4: Simultaneous Preference Control}
\label{sec:multi}

We extend SPT to two simultaneous preferences by combining release height and placement location, yielding four commands: low--center, low--corner, high--center, and high--corner. Source and target demonstrations cover all four combinations, but only source demonstrations have preference labels. The recognizer predicts both labels independently and achieves 100\% accuracy on each label and on the complete pair; therefore, inferred and ground-truth target labels coincide and we omit a separate Oracle GT condition.

Table~\ref{tab:multi} compares Source-only, Same-init plain FT, random labels, and SPT using the same two-axis Stage~A initialization, target demonstrations, and adaptation schedule. We evaluate 40 held-out scenes under all four commands, for 160 rollouts. Placement is evaluated from the released object's final XY position. Because all four combinations appear in training, this experiment evaluates simultaneous preference control rather than generalization to unseen combinations.


\begin{table}[t]
\caption{Two-preference target adaptation (\%). Preference following is reported for each preference and for both on the same rollout.}
\label{tab:multi}
\centering\small
\setlength{\tabcolsep}{3pt}
\begin{tabular}{@{}lccccc@{}}
\toprule
&\multicolumn{3}{c}{Preference Following}
&\multicolumn{1}{c}{\multirow{2}{*}{\shortstack[c]{Task\\Success}}}
&\multicolumn{1}{c}{\multirow{2}{*}{\shortstack[c]{Joint\\Success}}}\\
\cmidrule(lr){2-4}
Method & Release & Place & Both & &\\
\midrule
Source-only & 78.8 & 71.9 & 62.5 & 33.8 & 29.4\\
Random labels & 43.8 & 45.6 & 24.4 & 91.9 & 21.9\\
Same-init plain FT & 70.6 & 61.3 & 42.5 & 90.6 & 40.0\\
SPT (Ours) & \textbf{100.0} & \textbf{93.8} & \textbf{93.8} & 90.6 & \textbf{88.8}\\
\bottomrule
\end{tabular}
\vspace{-10pt}
\end{table}

SPT follows both preferences in 93.8\% of rollouts, compared with 42.5\% for Same-init plain FT and 24.4\% for random labels. Joint success is 88.8\%, versus 40.0\% and 21.9\%, while SPT and Same-init plain FT have the same 90.6\% task success. Thus, the improvement comes from following the requested execution rather than task completion.

\emph{Summary.} SPT supports two simultaneous execution preferences in this setting, with 93.8\% command following and 88.8\% joint success without target preference annotations.
\section{Conclusion and Discussion}
\label{sec:conclusion_discussion}
\subsection{Conclusion}
We studied preference transfer without target preference annotations. SPT jointly learns preference-conditioned action and preference recognition on labeled source tasks, then uses the recognizer to self-label target demonstrations for adaptation. Across five simulated axes, the recognizer achieves 96.4\% target accuracy and SPT improves preference following by 24.4 percentage points over Naive FT with similar task success. SPT also transfers preference control across task types on a real robot and supports two simultaneous preferences in simulation. Ablations show that effective target adaptation depends on reliable labels associated with the demonstrated behavior, suggesting that source preference annotations can be reused across target tasks.

\subsection{Limitations and Future Work}
The current formulation assumes a predefined preference vocabulary and target demonstrations that contain the relevant behaviors. Our experiments primarily study binary preferences, with simultaneous control evaluated on two axes whose combinations are observed during training. Extending SPT to continuous or open-vocabulary preferences, unseen preference combinations, and recognition with abstention for out-of-vocabulary behaviors are directions for future work.

\section*{Acknowledgment}
This work was supported in part by the National Science Foundation under Grant No. 2143435. We also thank the Coordinated Science Laboratory (CSL) at the University of Illinois Urbana-Champaign for providing research facilities to support this work.
\balance
\bibliographystyle{IEEEtran}
\bibliography{references}


\end{document}